\documentclass{article} 
\usepackage{iclr2016_conference,times,wrapfig}
\usepackage{url}
\usepackage{amsfonts}
\usepackage{subfigure}
\usepackage{graphicx}

\newcommand{\dist}[1]{\mathrm{Distr}\left[#1\right]}

\newcommand{\KL}[0]{\textrm{KL}} %
\newcommand{\X}[0]{\mathcal{X}}
\newcommand{\Y}[0]{\mathcal{Y}}
\newcommand{\D}[0]{\mathcal{D}}
\newcommand{\batchsize}[0]{\mathrm{batch size}}

\title{Towards Principled Unsupervised Learning}

\author{Ilya Sutskever$^1$, Rafal Jozefowicz$^1$, Karol Gregor$^2$, Danilo Rezende$^2$, Tim Lillicrap$^2$, Oriol Vinyals$^1$  \\
Google Brain$^1$ and Google DeepMind$^2$\\
\texttt{\{ilyasu,rafalj,karolg,danilor,countzero,vinyals\}@google.com} \\
}

\begin{document}

\maketitle
\begin{abstract}

General unsupervised learning is a long-standing conceptual problem in
machine learning.  Supervised learning is successful because it can be
solved by the minimization of the training error cost function.
Unsupervised learning is not as successful, because the unsupervised
objective may be unrelated to the supervised task of interest.  For an
example, density modelling and reconstruction have often been used for
unsupervised learning, but they did not produced the sought-after
performance gains, because they have no knowledge of the supervised
tasks.

In this paper, we present an unsupervised cost function which we name
the Output Distribution Matching (ODM) cost, which measures a divergence
between the distribution of predictions and distributions of labels.
The ODM cost is appealing because it is consistent with the supervised
cost in the following sense: a perfect supervised classifier is also
perfect according to the ODM cost.  Therefore, by aggressively
optimizing the ODM cost, we are almost guaranteed to improve our
supervised performance whenever the space of possible predictions is
exponentially large.

We demonstrate that the ODM cost works well on number of small and
semi-artificial datasets using no (or almost no) labelled training
cases.  Finally, we show that the ODM cost can be used for one-shot
domain adaptation, which allows the model to classify inputs that
differ from the input distribution in significant ways without the need
for prior exposure to the new domain. 

\end{abstract}

\section{Introduction}

Supervised learning is successful for two reasons: it is equivalent to
the minimization of the training error, and stochastic gradient
descent (SGD) is highly effective at minimizing training error.  As a
result, supervised learning is robust, reliable, and highly successful
in practical applications.

Unsupervised learning is not as successful, mainly because it is not
clear what the unsupervised cost function should be.  The goal of
unsupervised learning is often to improve the performance of a
supervised learning task for which we do not have a lot of data.  Due
to the lack of labelled examples, unsupervised cost functions do not
know which of the many possible supervised tasks we care about. As a
result, it is difficult for the unsupervised cost function to improve
the performance of the supervised cost function.  The disappointing
empirical performance of unsupervised learning supports this view.

In this paper, we present a cost function that generalizes the ideas
of \cite{cryptogram}.  We illustrate the idea in the setting of speech
recognition.  It is possible to evaluate the quality of a speech
recognition system by measuring the linguistic plausibility of its
typical outputs, without knowing whether these outputs are correct for
their inputs.  Thus, we can measure the performance of our system
without the use of any input-output examples.

We formalize and generalize this idea as follows:
in conventional supervised learning, we are trying to find an unknown
function $F$ from $\X$ to $\Y$. Each training case $(x_i,y_i)$ imposes
a soft constraint on $F$:
\begin{equation}
F(x_i)=y_i
\end{equation}
We solve these equations by local optimization with the
backpropagation algorithm \citep{backprop}.

We now present the unsupervised cost function.  Let $\D$ be the true
data distribution over the input-output pairs $(x,y) \sim \D$.  While
we do not have access to many labelled samples $(x,y) \sim \D$, we often
have access to large quantities of unlabelled samples from $x\sim
\D$ and $y \sim\D$.  This assumption is valid whenever unlabelled
inputs and unlabelled outputs are abundant.

We can use \emph{uncorrelated samples} from $x \sim \D$ and $y \sim
\D$ to impose a valid constraint on $F$:
\begin{equation}
\dist{F(x)} = \dist{y}
\end{equation}
where $\dist{z}$ denotes the distribution of the random variable $z$.
This constraint is valid in the following sense: if we have a
function $F$ such that $F(x_i) = y_i$ is true for every possible training
case, then $\dist{F(x)} = \dist{y}$ is satisfied as well.  Thus,
the unsupervised constraint can be seen as an additional labelled
``training case'' that conveys a lot of information whenever the output space is very large.

\begin{figure}[]
\centering
\subfigure{
  \includegraphics[scale=0.2]{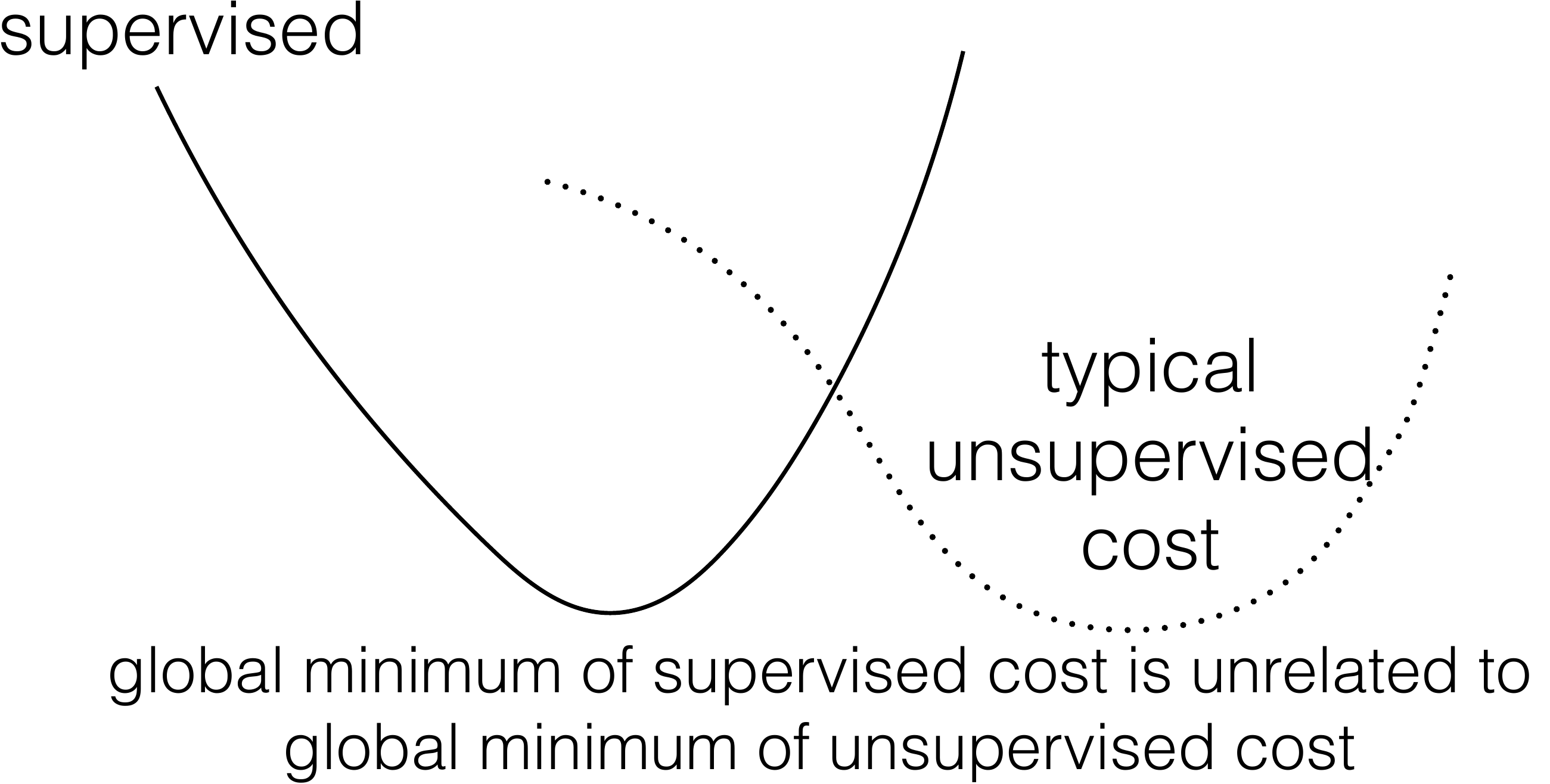}
  \hspace{1.5cm}
  \includegraphics[scale=0.2]{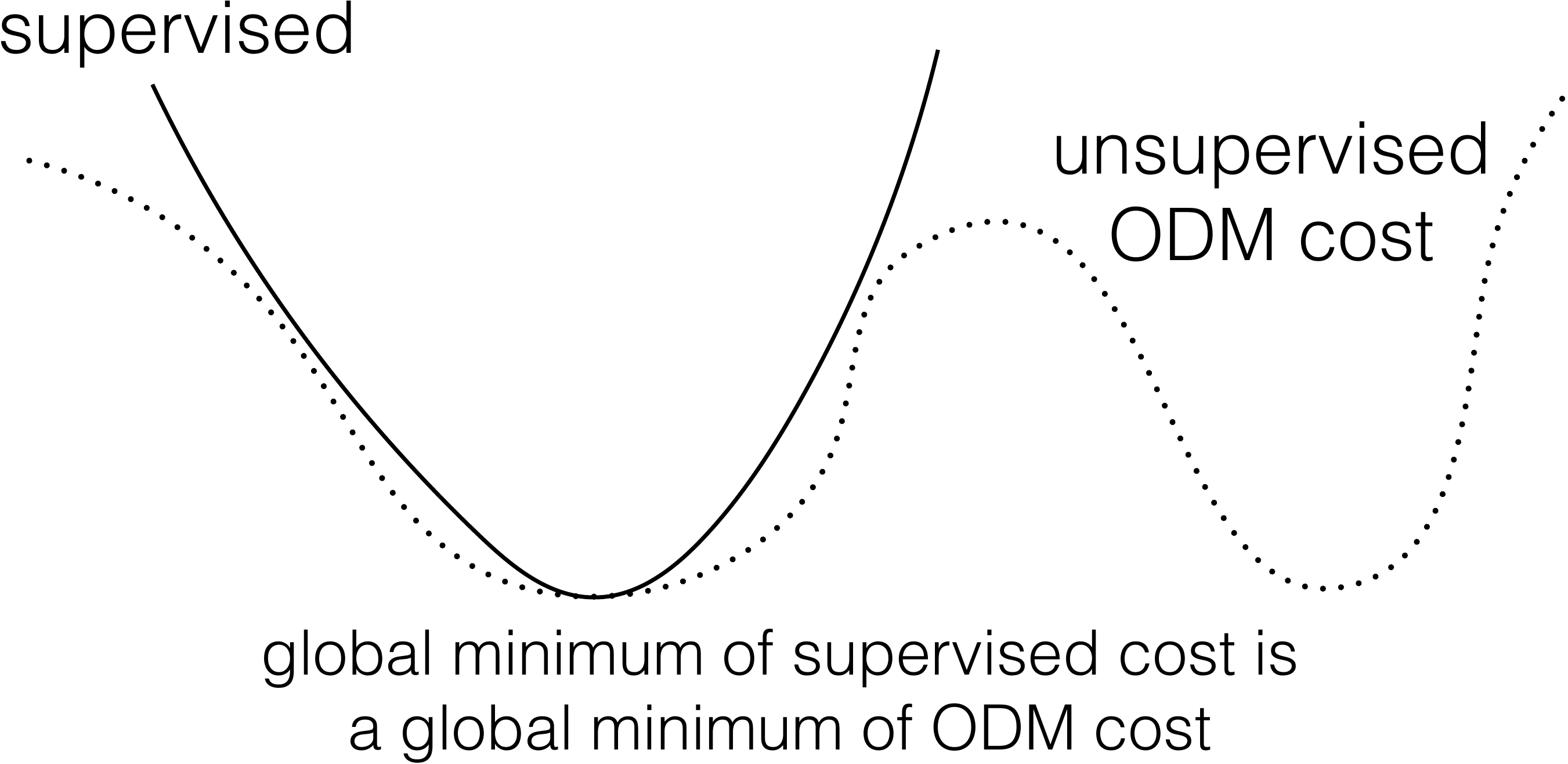}  
}
\caption{\small Existing unsupervised cost functions {\bf (left)}
  versus the ODM cost in the limit of infinite data {\bf(right)}.  The
  global minima of typical unsupervised cost functions, such as the
  reconstruction cost, are unrelated to the minima of the supervised
  cost.  The ODM cost. In contrast, the global minimum of the
  supervised cost function is also a global minimum of the ODM cost
  function whenever there is a function that can perfectly represent
  the mapping.  The implication is not true in reverse because the ODM
  cost can have additional global minima that are not global minima of
  the supervised cost.  This seemingly surprising property is possible
  because the ODM cost uses unlabelled examples from both the input
  domain and the output domain, while conventional unsupervised cost
  functions ignore the output domain.  }
\label{fig:diffcosts0}
\end{figure}

This constraint can be turned into the following cost function
\begin{equation}
\label{eqn:cost_kl}
\KL[\dist{y} \| \dist{F(x)}]
\end{equation}
We call it the output distribution matching (ODM) cost function, the cost
literally matches the distributions of the outputs. 

For the constraint $\dist {F(x)} = \dist{y}$ to be highly informative of 
the optimal parameters, it is necessary for the output space $\Y$ to be 
large,  since otherwise the ODM cost will be trivial to optimize.
If the output space is large, then the ODM cost is, in principle,
substantially more useful than conventional unsupervised cost
functions which are unrelated to the ultimate supervised cost
function.  In contrast, the ODM cost is nearly guaranteed to improve
the final supervised performance whenever there exists a very high
performing function, because a supervised function that perfectly maps
every input to its desired output is also a global minimum of the ODM
cost.  See Figure \ref{fig:diffcosts0}.  It also means that a
practitioner has a high chance of improving their supervised
performance if they are able to optimize the ODM cost.

We also show how the ODM cost can be used for one-shot domain
adaptation.  Given a test datapoint from a distribution that is very
different from the training one, our model can, in certain settings,
identify a nearly unique corresponding datapoint that happens to obey
the training data distribution, which can then be classified
correctly.  We implement this idea using a generative model as a
mechanism for aligning two distributions without supervision.

This allows each new test case to be from significantly different
distribution, so it is no longer necessary to assume that the test set
follows a particular distribution.  This capability allows our models,
in principle, to work robustly with unexpected data.  However, the
ability to perform one-shot domain adaptation is not universal, and it
can be achieved only in certain restricted settings.

\section{Related work}

There has been a number of publications that considered matching
statistics as an unsupervised learning signal.  The early work of
\cite{cryptogram} casts the problem of unsupervised OCR as a problem
of decoding a cipher, where there is an unknown mapping from images to
characters identities, which must be inferred from the known
statistics of language.  This idea has been further explored in the
context of machine translation (typically in the form of matching bigrams)
\citep{knight_et_al_2006_unsup,mccallum,snyder,knight_decipherment},
and in unsupervised lexicon induction
\citep{fung_and_mckeown,koehn_and_knight,liangHaghighi,mt}.  Notably,
\cite{knight_et_al_2006_unsup} discusses the connection between
unsupervised learning and language decipherment, and formulates a
generative model similar to the one presented in this work.

Similar ideas have recently been proposed for domain adaptation where
a model learns to map the new distribution back onto the training
distribution \citep{domain-confusion1,domain-confusion2}.  These
approaches are closely related to the ODM cost, since they are
concerned with transforming the new distribution back to the
training distribution. 

There has been a lot of other work on unsupervised learning with
neural networks, which is largely concerned with the concept of
``pre-training''.  In pre-training, we first train the model with an
unsupervised cost function, and finish training the model with the
supervised cost.  This concept was introduced by \cite{pretrain1} and
\cite{pretrain2} and later by \cite{pretrain4}.  Other work used the
$k$-means objective for pre-training \citep{coates}, and this list of
references is far from exhaustive.  These ideas have also been used
for semi-supervised learning \citep{kingma2014semi,rasmus2015semi} and
for transfer learning \citep{mesnil2012unsupervised}.

More recent examples of unsupervised pre-training are the Skip-gram
model \citep{skip-gram} and its generalization to sentences, the
Skip-thought vectors model of \cite{skip-thought}.  These models use
well-motivated unsupervised objective functions that appear to be
genuinely useful for a wide variety of language-processing tasks.

\section{Methods}

In this section, we present several approaches for optimizing the ODM cost.

\subsection{ODM Costs as Generative Models}
\label{sec:gen}

The ODM cost can be formulated as the following generative
model.  Let $P(x)$ be a model fitted to the marginal $x\sim \D$, and
let $P_\theta(y)$ be the distribution:
\begin{equation}
\label{eqn:gen1}
  P_\theta(y) = \sum_x P_\theta(y|x)P(x)
\end{equation}
The objective is to find a conditional $P_\theta(y|x)$ (which
corresponds to $F_\theta(x)$ where $\theta$ are the parameters) so
that $P_\theta(y)$ matches the marginal distribution $y\sim \D$.  If
$P(x)$ is an excellent model of $x\sim \D$, then the cost $E_{y\sim
  \D}[-\log P_\theta(y)]$ is precisely equivalent to the ODM cost of
Eq.~\ref{eqn:cost_kl}, modulo an additive constant.  A similar
generative model was presented by \cite{knight_et_al_2006_unsup}.

It is desirable to train generative models using the variational
autoencoder (VAE) of \cite{vae}.  However, VAE training forces $y$ to
be continuous, which is undesirable since many domains of interest are
discrete. We address this by proposing the following generative model,
which we term the $xyh$-model:
\begin{eqnarray}
  P_x(x) = \int_h P_x(x|h) P(h) \mathrm{d} h \\
  P_x(y) = \int_h P_y(y|h) P(h) \mathrm{d} h
\end{eqnarray}
whose cost function is
\begin{equation}
  L = E_{x\sim \D}[-\log P_x(x)] + E_{y\sim \D}[-\log P_y(y)]
\end{equation}
Here $h$ is continuous while $x$ and $y$ are either continuous or
discrete.  It is clear that this model can also be trained on labelled
$(x,y)$, whenever such data is available.

Although the negative log probability of the $xyh$-model is not
identical to the ODM cost, the two are closely related. Specifically,
whenever the capacity of $P_x$ and $P_y$ is limited, the $xyh$-model
will be forced to represent the structure that is common to $\dist{x}$
and $\dist{y}$ in $P(h)$, while placing the domain specific structures
into $P_x(x|y)$ and $P_y(x|y)$.  For example, in case of speech
recognition, $P(h)$ could contain the language model, since it is not
economical to store the language model in $P_x(x|h)$ and $P_y(y|h)$.

\begin{figure}[t!!]
\centering
\subfigure{
  \includegraphics[scale=0.2]{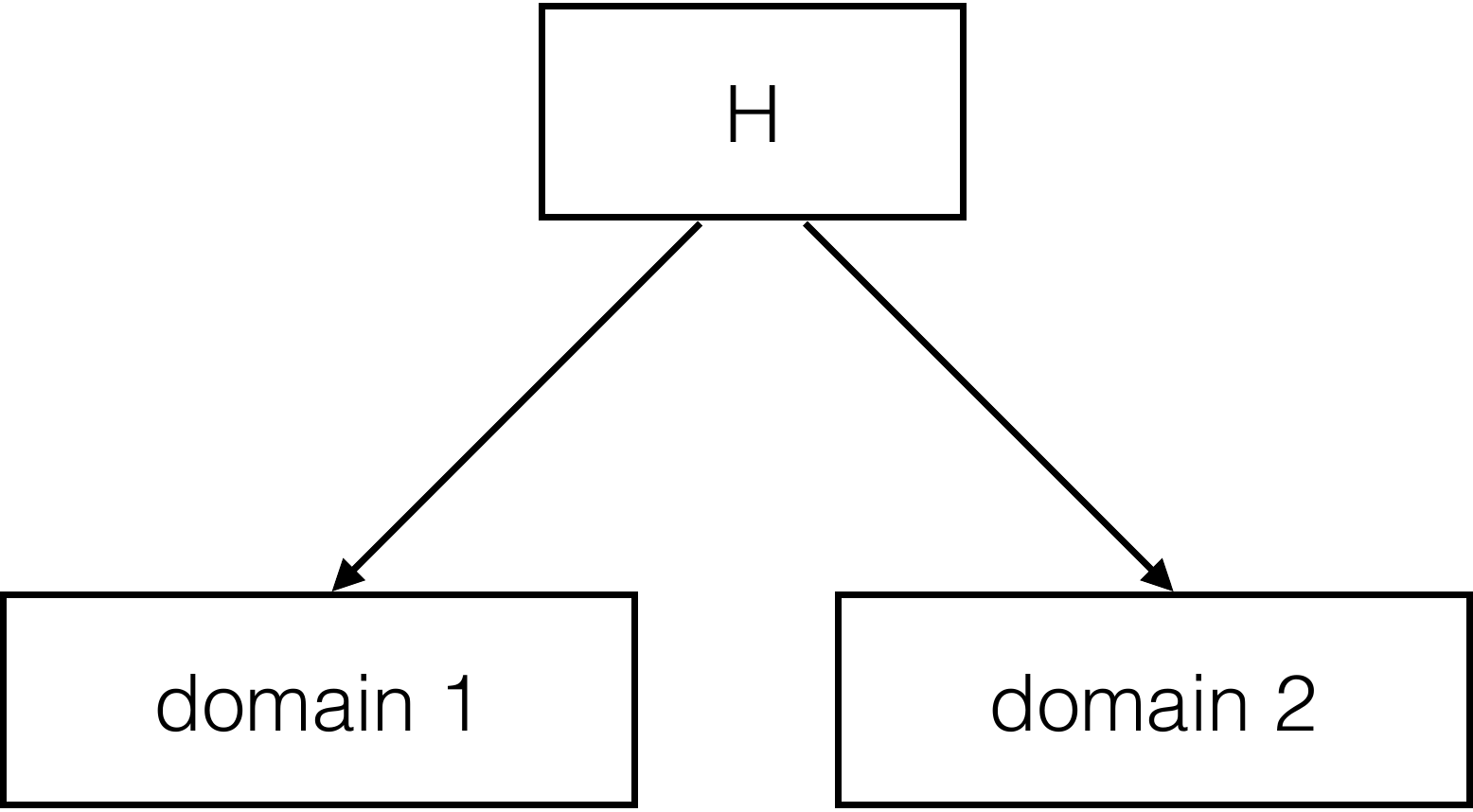}
  \hspace{1.5cm}
  \includegraphics[scale=0.2]{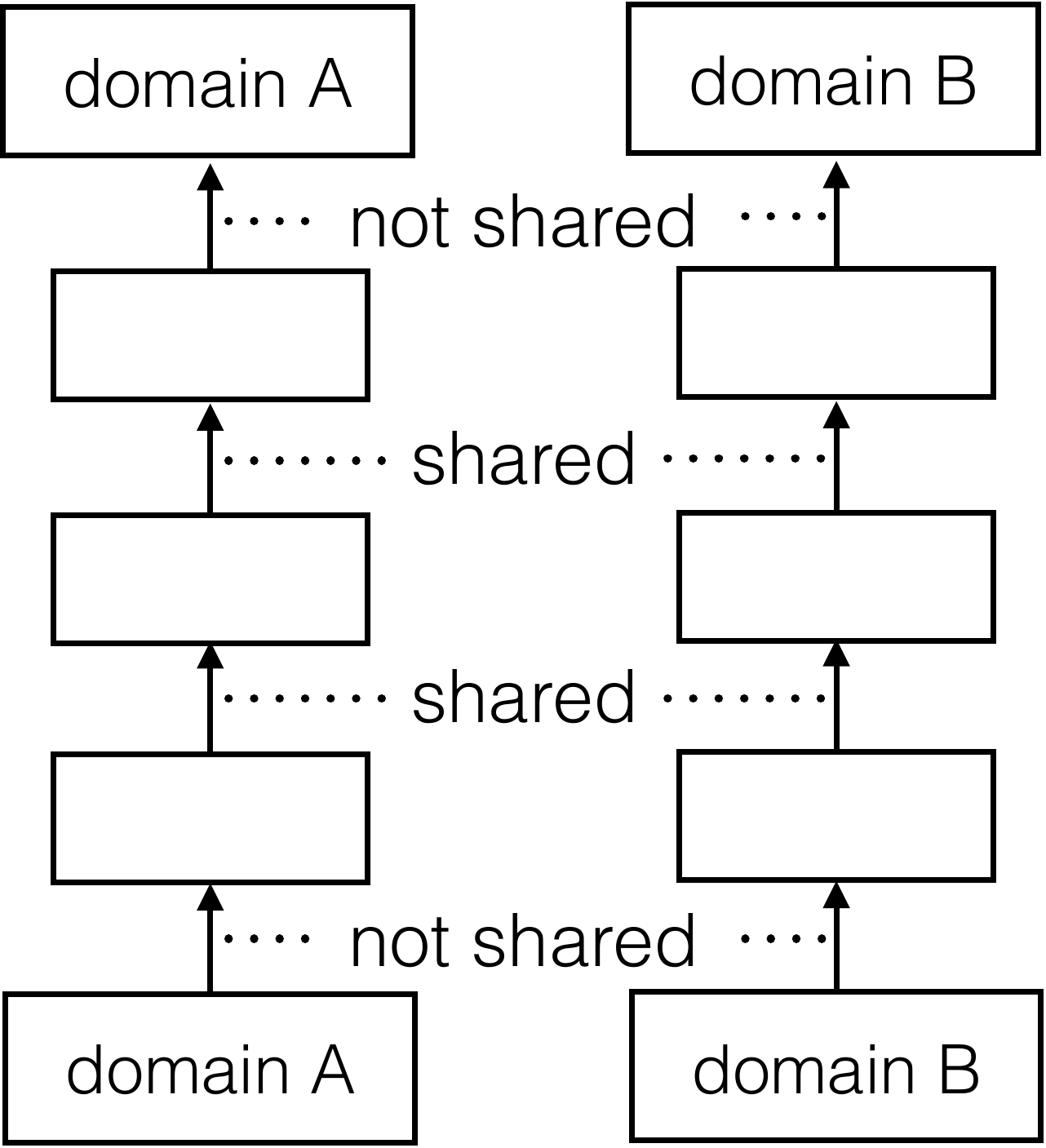}
  \hspace{1.5cm}
  \includegraphics[scale=0.2]{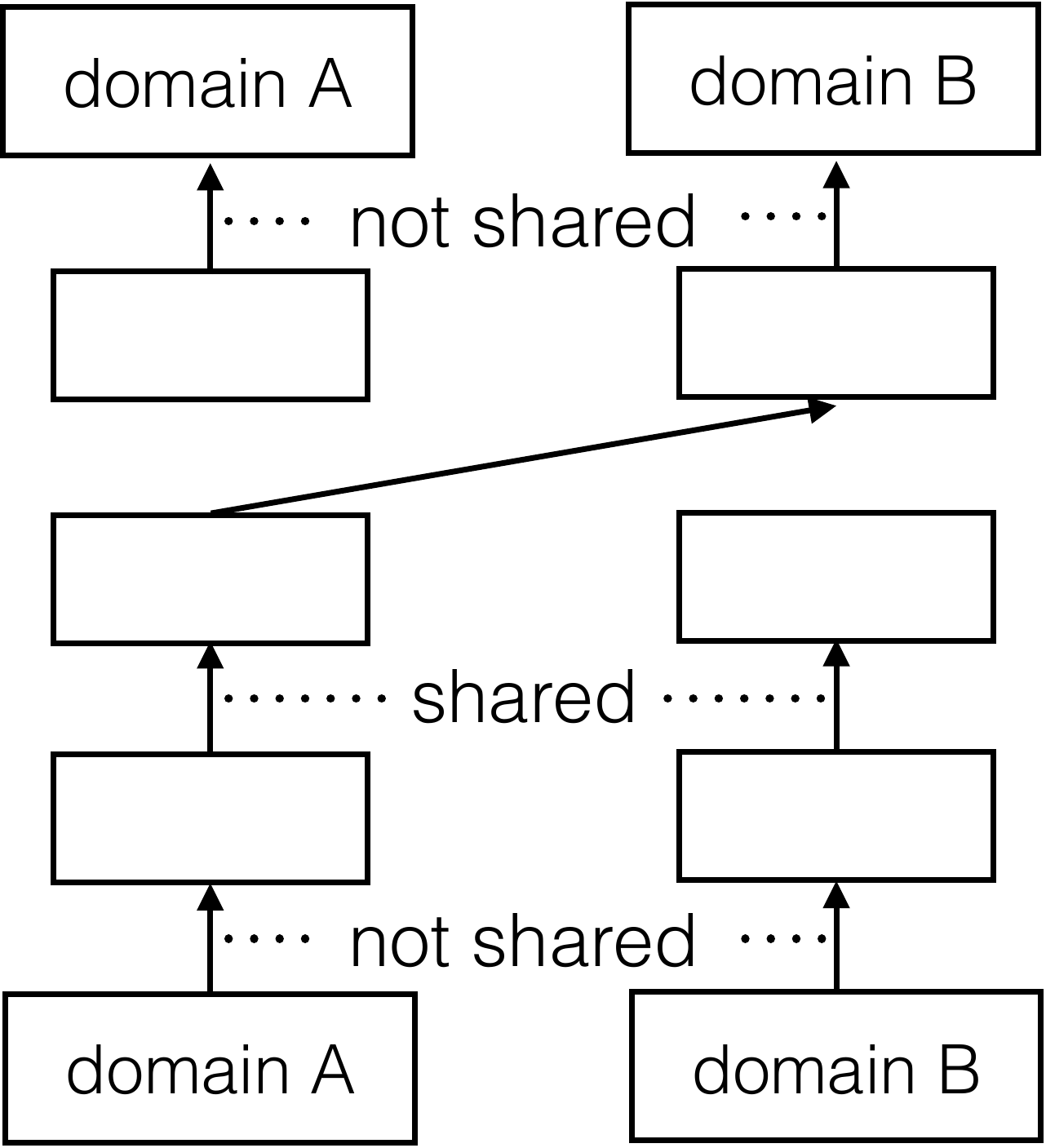}  
}
\caption{\small The dual autoencoder.  {\bf (Left:)} The
  $xyh$-generative model that motivated the dual autoencoder. {\bf
    (Middle:)} The dual autoencoder during training.  Note that it can
  be trained in an entirely unsupervised fashion on data from
  different domains.  The dual autonecoder learns a representation
  that is compatible with both domains, in a manner that is entirely
  unsupervised.  {\bf (Right:)} The dual autoencoder test time. It
  often successfully learns the correspondence between the domains
  even though it is not trained for this task.}
\label{fig:dualae}
\end{figure}

\subsection{The Dual Autoencoder}

We implemented the $xyh$-generative model but had difficulty getting
sensible results in our early experiments.  We were able to get better
results by designing an novel autoencoder model that is inspired by
the $xyh$-model, which we call the dual autoencoder,
which is shown in Figure \ref{fig:dualae}.  It consists of two
autoencoders whose ``innermost weights'' are shared with each other.
More formally, the dual autoencoder has the form
\begin{eqnarray}
x' = f(A_0 f(W_N f(W_{N-1} \ldots f(W_1 f(B_0 x))\ldots ))) \label{eqn:ae1} \\
y' = f(A_1 f(W_N f(W_{N-1} \ldots f(W_1 f(B_1 y))\ldots ))) \label{eqn:ae2} 
\end{eqnarray}
where $f$ is the nonlinearity, and the cost is
\begin{equation}
L = E_{x\sim \D}\left[L_1(x, x')\right] + E_{y\sim \D}\left[L_2(y, y')\right]
\end{equation}
where $L_1$ and $L_2$ are appropriate loss functions. 

Eqs.~\ref{eqn:ae1} and \ref{eqn:ae2} describe two autoencoders that
map $x$ to $x'$ and $y$ to $y'$, respectively.  By sharing the weights
$W_1\ldots,W_N$ between the autoencoders, the matrices $(A_0,B_0)$ and
$(A_1,B_1)$ are encouraged to use compatible representations for the
two modalities that align $x$ with $y$ in the absence of a direct
supervised signal.  While not principled, we found this approach to
work surprisingly well on simple problems.

\subsection{Generative Adversarial Networks Training}

The Generative Adversarial Network \citep{gan} is a procedure for
training a ``generator'' to produce samples that are statistically
indistinguishable from a desired distribution.  The generator is a
neural network $G$ that transforms a source of noise $z$ into samples
from some distribution:
\begin{equation}
z \to G(z)  
\end{equation}

The GAN training algorithm maintains an adversary $D(z) \to[0,1]$
whose goal is to distinguish between samples $x$ from the data
distribution and samples from the generator $G(z)$, and the generator
$G$ learns to fool the discriminator.  Eventually, if GAN training is
successful, $G$ converges to a model such that the distribution of
$G(z)$ is indistinguishable from the target distribution.

The generative adversarial network offers a direct way of training
generative models, and it had enjoyed considerable success in learning
models of natural images \citep{denton}.  We use the GAN training
method to train our unsupervised objective $\dist{F(x)} = \dist{y}$ by
requiring that $F$ produces samples that are indistinguishable from
the target distribution.

\section{Experiments}

\subsection{Dual Autoencoders on MNIST Permutation Task}

We begin exploring the ODM cost by selecting a simple artificial task
where the distributions over $\X$ and $\Y$ have rich internal
structure but whose relationship is simple.  We chose $\Y$ to be the
distribution of MNIST digits, and $\X$ to be the distribution of MNIST
digits whose pixels are permuted (with the same permutation on all
digits). See Figure \ref{fig:dualae}.  We call it the \emph{MNIST
  permutation task}.  The goal of the task is to learn the unknown
permutation using no supervised data.

We used a dual autoencoder whose architecture is 784-100-100-100-784,
where the weights in the 100-100-100 subnetwork were shared between
the two autoencoders.  While the results were insensitive to the
learning rate, it was important to use a random initialization that is
substantially smaller than is typically prescribed for training neural networks
models (namely, a unit Gaussian scaled by $0.003 /
\sqrt{\#\mathrm{rows}+\#\mathrm{cols}}$ for each matrix).

\begin{wrapfigure}[28]{r}{0.4\textwidth}
\centering
\includegraphics[scale=0.35]{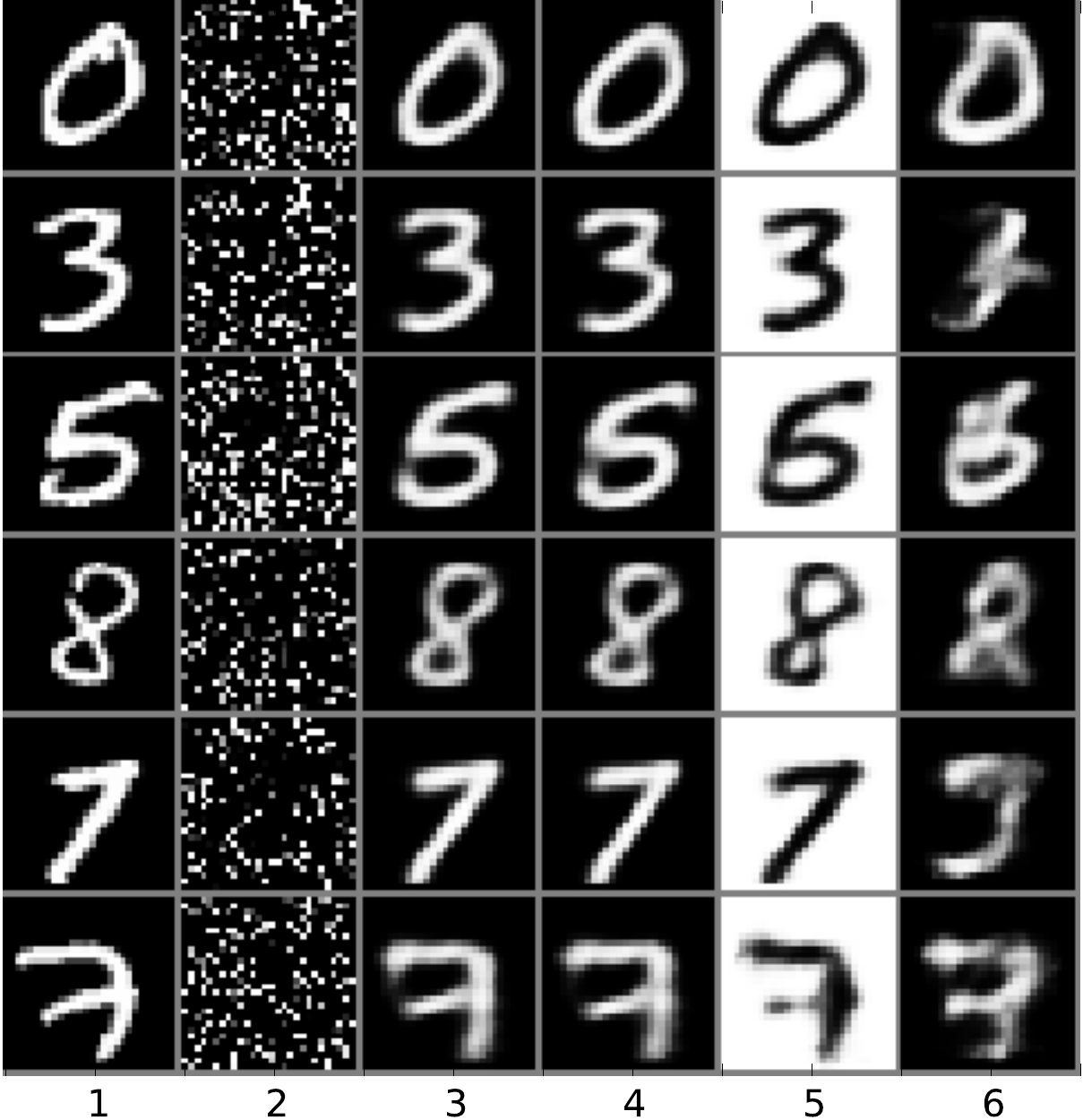}  
 \caption{\small Illustrative performance of the various models on the
   MNIST permutation task.  Column 1: data.  Column 2: the permuted
   data. Column 3: {\bf Supervised} Autoencoder; Column 4: {\bf
     Unsupervised} sigmoid dual autoencoder; Column 5: {\bf Unsupervised}
   tanh dual autoencoder; Column 6: {\bf Unsupervised} relu Dual
   Autoencoder.  Notice that the dual autoencoder that
   used the Tanh nonlinearity was able to find the correct
   correspondence while inverting the images.  We do not understand
   why this happens.  The sigmoid dual autoencoder is by far the best
   method for recovering a mapping between two corresponding datasets.
 }
\label{fig:perm}
\end{wrapfigure}

We found that the dual autoencoder was easily able to recover the
permutation without using any input-output examples, as shown in
Figure \ref{fig:perm}.

\setlength{\columnsep}{15pt}%

Curiously, we found that the sigmoid nonlinearity achieved by far the
best results on this task, and that the Relu and the Tanh nonlinearity
were much less successful for a wide range of hyperparameters.  Paradoxically,
we suspect that the slower learning of the sigmoid nonlinearity was beneficial for
the dual autoencoder, because it caused the ``capacity'' of the
100-100-100 network to grow slowly; it is plausible that the most
critical learning stage took place when the 100-100-100 network had
the lowest capacity.

\subsubsection{Dual Autoencoder for Ciphers}

We also tested the dual autoencoder on a character and a word cipher,
tasks that were also considered by \cite{knight_et_al_2006_unsup}.  In
this task, we are given two text corpora that follow the same
distribution, but where the characters (or the words) of one of the
corpora is scrambled.  In detail, we convert a text file into a list
of integers by randomly assigning characters (or words) to integers
and by consistently using this assignment throughout the file.  By
doing this twice, we get two data sources that have identical
underlying statistics but different symbols.  The goal is to find the
hidden correspondence between the symbols in both streams.  It is not
a difficult task since it can be accomplished by inspecting the
frequency. Despite its simplicity, this task provides us with another
way to evaluate our model.

We used the dual autoencoder architecture as before, where the input
(as well as the the desired output) is represented with a bag of 10
consecutive characters from a random sentence from the text corpus.
The loss function is the softmax cross entropy.

For the character-level experiments, we used the architecture
100-25-25-100, and for the word-level experiments we used
1000-100-100-1000 --- where we used a vocabulary of the 1000 most
frequent symbols in both data streams.  Both architectures achieved
perfect identification of the symbol mapping in less than 10,000
parameter updates when the sigmoid nonlinearity was used.

While the dual autoencoder was successful on these tasks, its success
critically relied on the underlying statistics of the two datasets
being very similar.  When we trained each autoencoder on data from a
different languages (namely on English and Spanish books from Project
Gutenberg), it failed to learn the correct correspondence between
words with similar meanings.  This indicates that the dual
autoencoder is sensitive to systematic differences in the two distributions.

\begin{figure}[]
\centering
\includegraphics[scale=0.25]{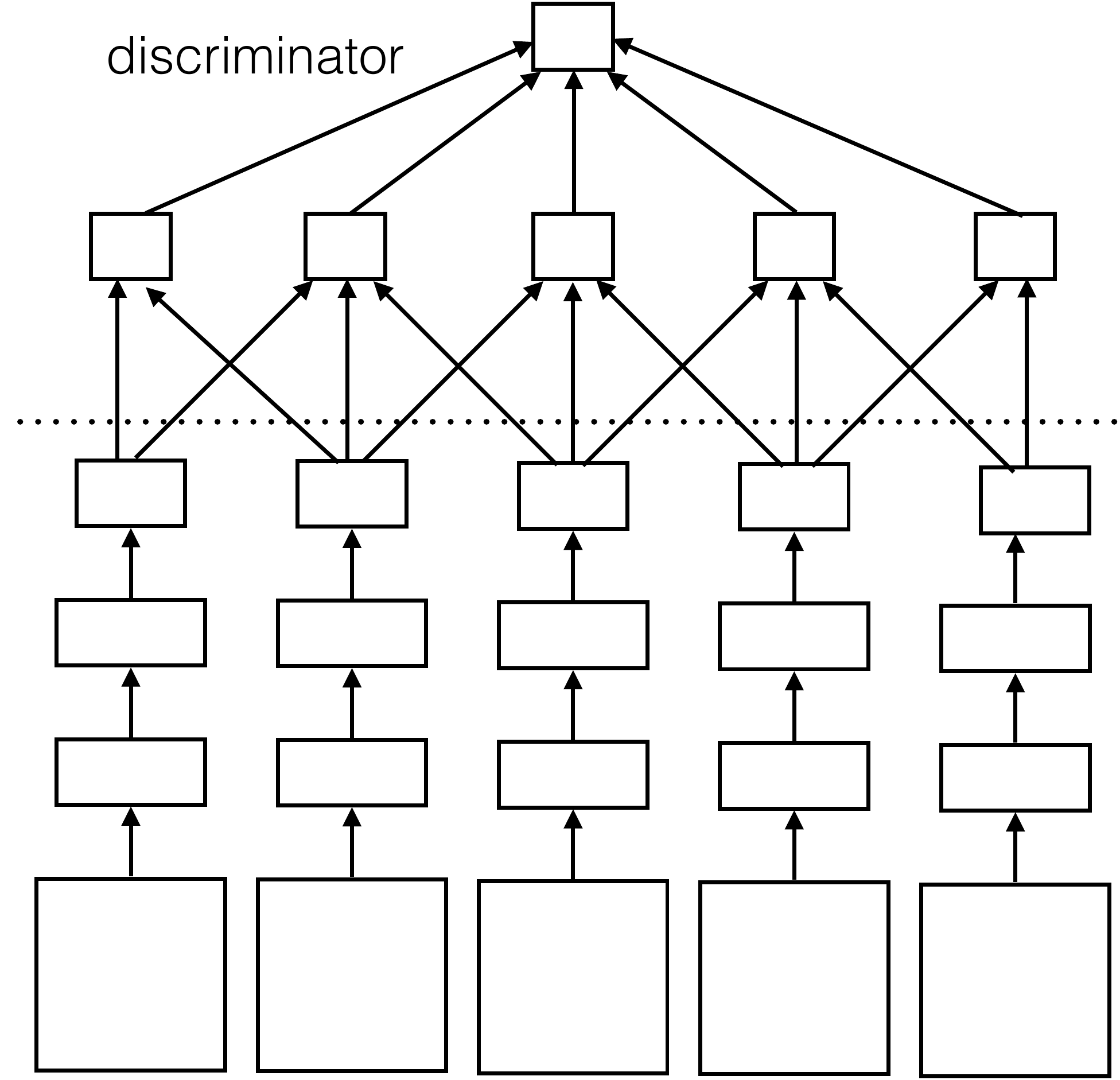}
\caption{\small An illustration of the MNIST classification setup.  The classifier
  is shown below the dotted line;  the discriminator is shown above it.  The first
  few layers of the discriminator are convolutional; the remaining layers are fully
  connected.
}
\label{fig:mnist_class}
\end{figure}

\subsection{Generative Adversarial Networks for MNIST classification}

Next, we evaluated the GAN-based model on an artificial OCR task that
was constructed from the MNIST dataset.  In this task, we used the
text from \emph{The Origin of Species} downloaded from Project
Gutenberg, and arbitrarily replaced each letter with a digit between 0
and 9 in a consistent manner.  We call this sequence of numbers the
\emph{MNIST label file}.  We then replaced each label with a random
image of an MNIST digit of the same class, thus obtaining a very long
sequence of images of MNIST digits which we call the \emph{MNIST image
  sequence}.

The goal of this problem was to train an MLP $F$ that maps MNIST
digits to 10-dimensional vectors without using any input-output
examples.  Instead, we train the classifier to match the 20-gram
statistics of the Origin of Species, which requires no input-output
examples.

We trained an MLP $F$ to map each MNIST digit into a 10-dimensional
vector representing their classification.  We used generative
adversarial training where the adversary is a 1-d CNN to ensure sure
that the distribution of 20 consecutive predictions
$(F(x_1),\ldots,F(x_{20}))$ is statistically indistinguishable from
the distribution of 20 consecutive labels from the MNIST label file
$(y_1,\ldots,y_{20})$, where the inputs $(x_1,\ldots,x_{20})$ are
randomly drawn from the MNIST image sequence.

A typical architecture of our classifier network was 784-300-300-10.
The adversary consist of a 1D convolutional neural network with the
following architecture: the first three layers are convolutional with
a kernel of width 7, whit the following numbers of filter maps:
10-200-200. It is followed by global max pooling over time, and the
remainder of the architecture consist of the fully connected layers
200-200-1.  The nonlinearity was always ReLU, except for the output
layer of both networks, which was the identity.

In these experiments, we used the squared loss and not the
cross-entropy loss for the supervised objective.  If we used the
softmax loss, our model's output layer would have to use the softmax
layer, which we found to not work well with GAN training.  The squared
loss was necessary for using a linear layer for the outputs.

It is notable that GAN training was also fragile, and the sensitive hyperparameters
are given here:
\begin{itemize}
\item We initialized each matrix of the generator with a unit Gaussian
  scaled by $1.4/\sqrt{\#\mathrm{rows} + \#\mathrm{cols}}$; for the
  discriminator we used $1.0/\sqrt{\#\mathrm{rows} + \#\mathrm{cols}}$
\item The $\batchsize$ is 200
\item Total number of parameter updates: 6000
\item Learning rate of generator:  $0.1 / \batchsize$ for 2000 updates; $0.03 / \batchsize$ for another 2000 updates; and then $0.01 / \batchsize$
  for 2000 updates
\item Learning rate of discriminator: $0.005 / \batchsize$
\item Learning rate of the supervised squared loss: $0.005 / \batchsize$
\end{itemize}
While learning was extremely rapid, it was highly sensitive to the choice
of learning rates.

While GAN training of the ODM cost alone was insufficient to find an
accurate classifier the problem, we were able to achieve \emph{{\bf
    4.7}}\% test error on MNIST classification using \emph{{\bf 4}}
labelled MNIST examples.  Here, each labelled example consists of a
\emph{single} labelled digit, and not a single sequence of 20 labelled
digits. The result is insensitive to the specific set of 4 labelled
examples.  When using \emph{2} labelled MNIST digits, the
classification test error increased to \emph{72}\%.

\section{One-shot Learning and Domain Adaptation}

If the function that we wish to learn has a small number of
parameters, then it should be possible to infer it from a very small
number of examples --- and in some cases, from a single example. 

Suppose that the training distribution is $\D$, but we are provided
with a single test sample $y\sim\D'$ where $\D'$ is an unknown test
distribution.  If $y$ contains enough information to uniquely identify
a simple function $G$ that maps samples $y\sim \D'$ to samples
$x\sim\D$, we will be able to classify samples $y\sim\D'$ without any
extra training: we could simply apply our existing classifier to
$G(y)$.  By inferring the function $G$ from scratch for each test
sample from scratch, the model becomes capable at correctly
classifying test cases whose statistics are completely different from the
training distribution, as well test cases that are as outliers
that would have otherwise confused or misled the classifier.  This
has the potential to make our classifier significantly more robust.

\setlength{\columnsep}{13pt}%
\begin{wrapfigure}[28]{r}{0.37\textwidth}
\centering
\includegraphics[scale=0.3]{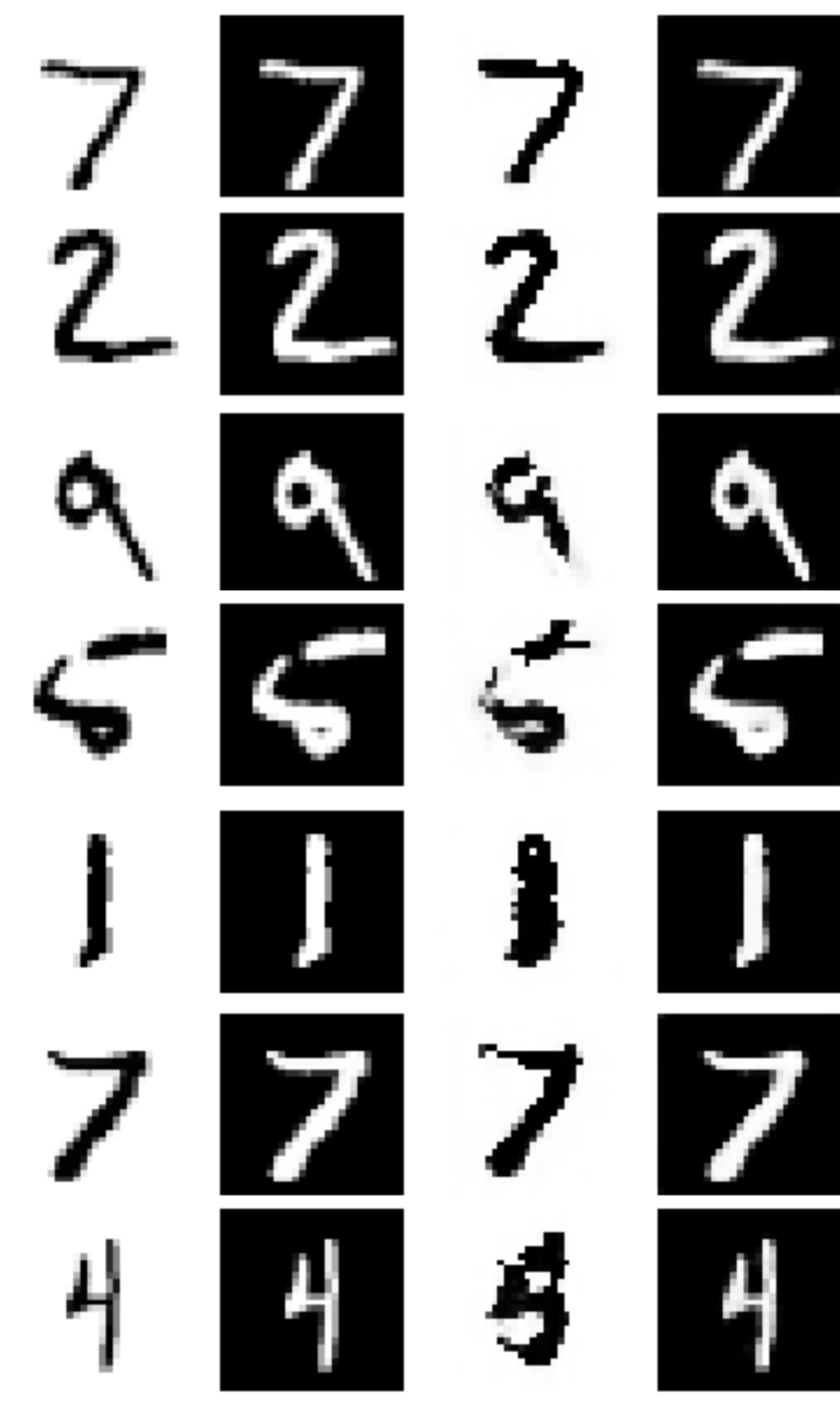}  
 \caption{\small An illustration of one-shot domain adaptation. Column
   1: sample original data; Column 2; 1-data ($y$); Column 3: the inferred
   $x$; Column 4: the reconstructed $y$.  The bottom row is an example
   of an incorrect reconstruction.  About 5\% of the test digits were
   reconstructed incorrectly.  Note that we infer the CNN $p(y|x)$ for 
   each test case separately.}
\label{fig:domainadapt}
\end{wrapfigure}

We propose to solve these domain adaptation tasks using the generative
model of Eq.~\ref{eqn:gen1}.  Let $P(x)$ be a model of the data
distribution and let $P_\theta(y|x)$ be an unknown likelihood
function.  Then, given a test case $y$ from a distribution that
differs from the data distribution, we can try to infer the unknown
$x$ by solving the following optimization problem:
\begin{equation}
  x^* = \mathrm{argmax}_{x,\theta} P_\theta(y|x)P(x)
\end{equation}
We emphasize that this optimization is run \emph{from scratch} for
each new test case $y$.  This approach can succeed only when the
likelihood $P_\theta(y|x)$ has few parameters that can be uniquely
determined from the above equation using just one sample.  The
conditions under which this is likely to hold are discussed in
sec.~\ref{sec:disc}.

We validate these ideas on use the MNIST dataset for training and
the 1-MNIST dataset for testing.  The 1-MNIST dataset is obtained by
replacing the intensity $u$ of each pixel with $1-u$. Thus the 1-MNIST
distribution differs very significantly from the MNIST distribution,
and as a result, neural networks trained on MNIST are incapable of
correctly classifying instances of 1-MNIST.

Our concrete model choices are the following: $P(x)$ is implemented
with a next-row-prediction LSTM with three hidden layers that has been
trained to fit the MNIST distribution with the binary cross entropy
loss, and $P(y|x)$ is a small convolutional neural network (CNN) with
one hidden layer:  its first convolution has 5 filters of size
5$\times$5 and its second convolution has one filter of size
5$\times$5.

Given a test image $y$ from the 1-MNIST dataset, we optimize $\log
P_\theta(y|x)P(x)$ over $x$ and $\theta$.  We used significant L2
regularization, and optimized this cost with $10^4$ steps of Adagrad
with 5 random restarts.  The results are illustrated in
Fig.~\ref{fig:domainadapt}.

While the results obtained in this section are preliminary, they show
that in addition to synthesis and denoising, generative models can be
used for aligning distributions and for expanding the set of
distributions to which the model is applied.  

\section{Discussion}

\subsection{When is Supervision Needed?}
\label{sec:disc}

When does the ODM fully determine the best $F$?  If the function class
is too capable, then $F$ can map any distribution input distribution
to any output distribution, which means that the ODM cost cannot
possibly recover $F$.  However, since the ODM cost is consistent, it
is likely to improve generalization by eliminating unsuitable
functions from consideration.  Further, if the output space $\Y$ is
small, then the ODM cost will not convey enough information to be of
significant help to the supervised cost.

\begin{figure}[t!]
\centerline{
  \includegraphics[width=0.3\linewidth]{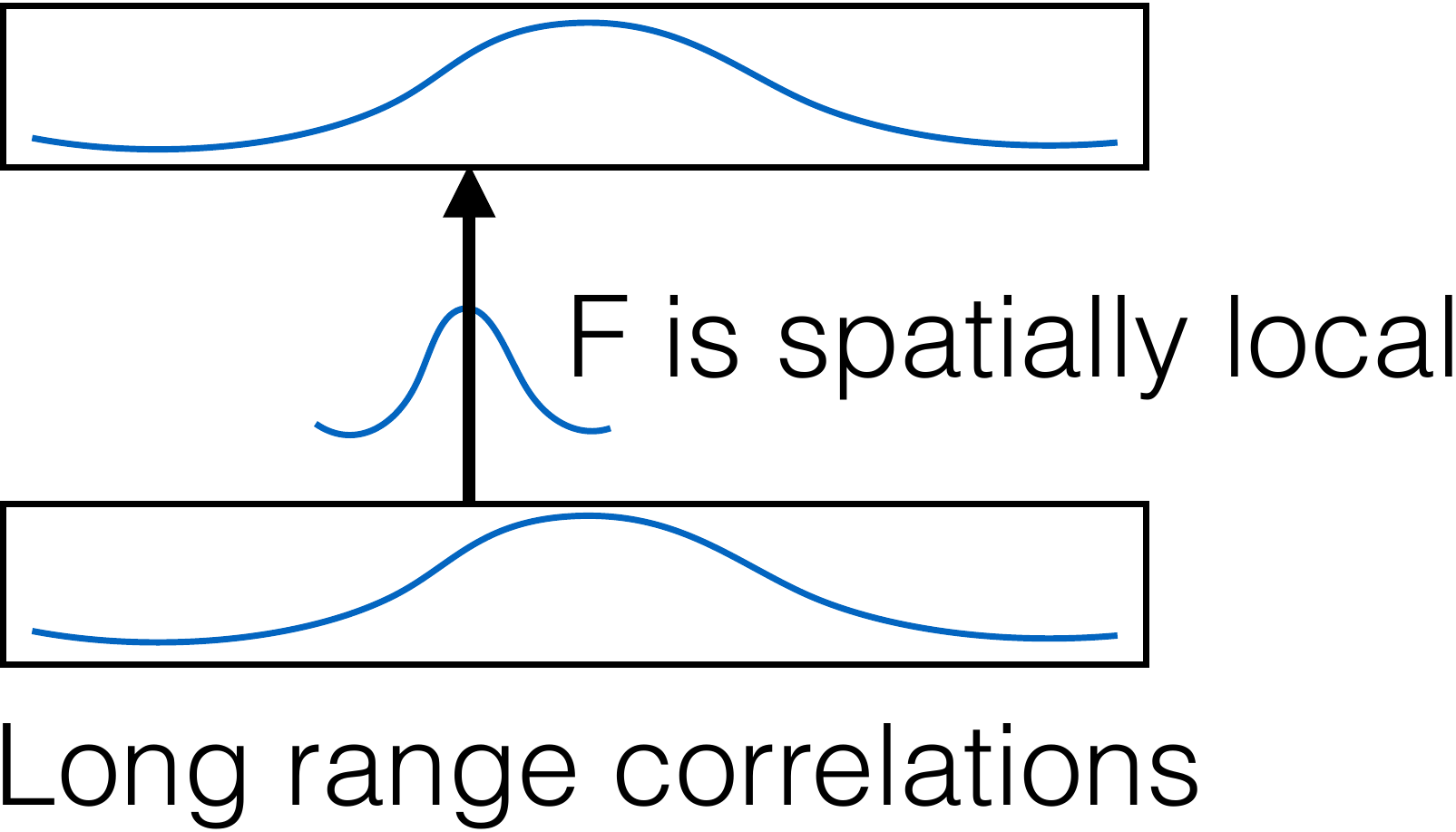}
}
\caption{\small An setting where the ODM function can recover the true $F$
  without help from a supervised objective.  If $F$ is incapable of
  modifying the long-range structure of the signal, then the ODM
  objective is likely to fully determine the best performing
  function $F$. }
\label{fig:situation}
\end{figure}

However, there is a setting where the ODM cost function could, in
principle, completely determine the best supervised function $F$.  It
should succeed whenever the input distribution and the output
distribution contains long-range dependencies, while $F$ is inherently
incapable of modifying the long range structure by being ``local'':
for example, if the input is a sequence, and we change the input in
one timestep, then the function's output will also change in a small
number of timestep.  This setting is illustrated in
Fig.~\ref{fig:situation}.   

\subsection{Limitations}

ODM-based training has several limitations.  The output space must be
large, and the ``shared hidden structure'' of the two space have to be
sufficiently similar.  The simple techniques used in this paper are
unlikely to successfully learn a mapping between two spaces if their
shared structures are insufficiently similar.  It is conceivable that
as we develop better methods for training generative models, it will
become possible to benefit from optimizing the ODM cost in a wider range
of problems.

\section{Conclusions}

In this paper, we showed that the ODM cost provides a generic approach
for unsupervised learning that is entirely consistent with the
supervised cost function.  Although we were not able to develop a
reliable method that can train, e.g., an attention model
\cite{bahdanau} with the ODM objective, we presented evidence that the
ODM objective provides a sensible way of training functions without
the use of input-output examples.  We expect better techniques for
optimizing the ODM cost to make it universally applicable and useful.

\small
\bibliography{unsup}

\begin{thebibliography}{25}
\providecommand{\natexlab}[1]{#1}
\providecommand{\url}[1]{\texttt{#1}}
\expandafter\ifx\csname urlstyle\endcsname\relax
  \providecommand{\doi}[1]{doi: #1}\else
  \providecommand{\doi}{doi: \begingroup \urlstyle{rm}\Url}\fi

\bibitem[Bahdanau et~al.(2014)Bahdanau, Cho, and Bengio]{bahdanau}
Bahdanau, Dzmitry, Cho, Kyunghyun, and Bengio, Yoshua.
\newblock Neural machine translation by jointly learning to align and
  translate.
\newblock \emph{arXiv preprint arXiv:1409.0473}, 2014.

\bibitem[Bengio et~al.(2007)Bengio, Lamblin, Popovici, Larochelle,
  et~al.]{pretrain4}
Bengio, Yoshua, Lamblin, Pascal, Popovici, Dan, Larochelle, Hugo, et~al.
\newblock Greedy layer-wise training of deep networks.
\newblock \emph{Advances in neural information processing systems},
  19:\penalty0 153, 2007.

\bibitem[Casey(1986)]{cryptogram}
Casey, Richard~G.
\newblock \emph{Text OCR by solving a cryptogram}.
\newblock International Business Machines Incorporated, Thomas J. Watson
  Research Center, 1986.

\bibitem[Coates et~al.(2011)Coates, Ng, and Lee]{coates}
Coates, Adam, Ng, Andrew~Y, and Lee, Honglak.
\newblock An analysis of single-layer networks in unsupervised feature
  learning.
\newblock In \emph{International conference on artificial intelligence and
  statistics}, pp.\  215--223, 2011.

\bibitem[Denton et~al.(2015)Denton, Chintala, Szlam, and Fergus]{denton}
Denton, Emily, Chintala, Soumith, Szlam, Arthur, and Fergus, Rob.
\newblock Deep generative image models using a laplacian pyramid of adversarial
  networks.
\newblock \emph{arXiv preprint arXiv:1506.05751}, 2015.

\bibitem[Fung \& McKeown(1997)Fung and McKeown]{fung_and_mckeown}
Fung, Pascale and McKeown, Kathleen.
\newblock A technical word-and term-translation aid using noisy parallel
  corpora across language groups.
\newblock \emph{Machine translation}, 12\penalty0 (1-2):\penalty0 53--87, 1997.

\bibitem[Gani et~al.(2015)Gani, Ustinova, Ajakan, Germain, Larochelle,
  Laviolette, Marchand, and Lempitsky]{domain-confusion2}
Gani, Yaroslav, Ustinova, Evgeniya, Ajakan, Hana, Germain, Pascal, Larochelle,
  Hugo, Laviolette, Fran{\c{c}}ois, Marchand, Mario, and Lempitsky, Victor.
\newblock Domain-adversarial training of neural networks.
\newblock \emph{arXiv preprint arXiv:1505.07818}, 2015.

\bibitem[Goodfellow et~al.(2014)Goodfellow, Pouget-Abadie, Mirza, Xu,
  Warde-Farley, Ozair, Courville, and Bengio]{gan}
Goodfellow, Ian, Pouget-Abadie, Jean, Mirza, Mehdi, Xu, Bing, Warde-Farley,
  David, Ozair, Sherjil, Courville, Aaron, and Bengio, Yoshua.
\newblock Generative adversarial nets.
\newblock In \emph{Advances in Neural Information Processing Systems}, pp.\
  2672--2680, 2014.

\bibitem[Haghighi et~al.(2008)Haghighi, Liang, Berg-Kirkpatrick, and
  Klein]{liangHaghighi}
Haghighi, Aria, Liang, Percy, Berg-Kirkpatrick, Taylor, and Klein, Dan.
\newblock Learning bilingual lexicons from monolingual corpora.
\newblock In \emph{ACL}, volume 2008, pp.\  771--779, 2008.

\bibitem[Hinton \& Salakhutdinov(2006)Hinton and Salakhutdinov]{pretrain2}
Hinton, Geoffrey~E and Salakhutdinov, Ruslan~R.
\newblock Reducing the dimensionality of data with neural networks.
\newblock \emph{Science}, 313\penalty0 (5786):\penalty0 504--507, 2006.

\bibitem[Hinton et~al.(2006)Hinton, Osindero, and Teh]{pretrain1}
Hinton, Geoffrey~E, Osindero, Simon, and Teh, Yee-Whye.
\newblock A fast learning algorithm for deep belief nets.
\newblock \emph{Neural computation}, 18\penalty0 (7):\penalty0 1527--1554,
  2006.

\bibitem[Huang et~al.(2006)Huang, Learned-Miller, and McCallum]{mccallum}
Huang, Gary, Learned-Miller, Erik~G, and McCallum, Andrew.
\newblock Cryptogram decoding for optical character recognition.
\newblock \emph{University of Massachusetts-Amherst Technical Report},
  6\penalty0 (45), 2006.

\bibitem[Kingma \& Welling(2013)Kingma and Welling]{vae}
Kingma, Diederik~P and Welling, Max.
\newblock Auto-encoding variational bayes.
\newblock \emph{arXiv preprint arXiv:1312.6114}, 2013.

\bibitem[Kingma et~al.(2014)Kingma, Mohamed, Rezende, and
  Welling]{kingma2014semi}
Kingma, Diederik~P, Mohamed, Shakir, Rezende, Danilo~Jimenez, and Welling, Max.
\newblock Semi-supervised learning with deep generative models.
\newblock In \emph{Advances in Neural Information Processing Systems}, pp.\
  3581--3589, 2014.

\bibitem[Kiros et~al.(2015)Kiros, Zhu, Salakhutdinov, Zemel, Torralba, Urtasun,
  and Fidler]{skip-thought}
Kiros, Ryan, Zhu, Yukun, Salakhutdinov, Ruslan, Zemel, Richard~S, Torralba,
  Antonio, Urtasun, Raquel, and Fidler, Sanja.
\newblock Skip-thought vectors.
\newblock \emph{arXiv preprint arXiv:1506.06726}, 2015.

\bibitem[Knight et~al.(2006)Knight, Nair, Rathod, and
  Yamada]{knight_et_al_2006_unsup}
Knight, Kevin, Nair, Anish, Rathod, Nishit, and Yamada, Kenji.
\newblock Unsupervised analysis for decipherment problems.
\newblock In \emph{Proceedings of the COLING/ACL on Main conference poster
  sessions}, pp.\  499--506. Association for Computational Linguistics, 2006.

\bibitem[Koehn \& Knight(2002)Koehn and Knight]{koehn_and_knight}
Koehn, Philipp and Knight, Kevin.
\newblock Learning a translation lexicon from monolingual corpora.
\newblock In \emph{Proceedings of the ACL-02 workshop on Unsupervised lexical
  acquisition-Volume 9}, pp.\  9--16. Association for Computational
  Linguistics, 2002.

\bibitem[Mesnil et~al.(2012)Mesnil, Dauphin, Glorot, Rifai, Bengio, Goodfellow,
  Lavoie, Muller, Desjardins, Warde-Farley, et~al.]{mesnil2012unsupervised}
Mesnil, Gr{\'e}goire, Dauphin, Yann, Glorot, Xavier, Rifai, Salah, Bengio,
  Yoshua, Goodfellow, Ian~J, Lavoie, Erick, Muller, Xavier, Desjardins,
  Guillaume, Warde-Farley, David, et~al.
\newblock Unsupervised and transfer learning challenge: a deep learning
  approach.
\newblock \emph{ICML Unsupervised and Transfer Learning}, 27:\penalty0 97--110,
  2012.

\bibitem[Mikolov et~al.(2013{\natexlab{a}})Mikolov, Chen, Corrado, and
  Dean]{skip-gram}
Mikolov, Tomas, Chen, Kai, Corrado, Greg, and Dean, Jeffrey.
\newblock Efficient estimation of word representations in vector space.
\newblock \emph{arXiv preprint arXiv:1301.3781}, 2013{\natexlab{a}}.

\bibitem[Mikolov et~al.(2013{\natexlab{b}})Mikolov, Le, and Sutskever]{mt}
Mikolov, Tomas, Le, Quoc~V, and Sutskever, Ilya.
\newblock Exploiting similarities among languages for machine translation.
\newblock \emph{arXiv preprint arXiv:1309.4168}, 2013{\natexlab{b}}.

\bibitem[Rasmus et~al.(2015)Rasmus, Valpola, Honkala, Berglund, and
  Raiko]{rasmus2015semi}
Rasmus, Antti, Valpola, Harri, Honkala, Mikko, Berglund, Mathias, and Raiko,
  Tapani.
\newblock Semi-supervised learning with ladder network.
\newblock \emph{arXiv preprint arXiv:1507.02672}, 2015.

\bibitem[Ravi \& Knight(2008)Ravi and Knight]{knight_decipherment}
Ravi, Sujith and Knight, Kevin.
\newblock Attacking decipherment problems optimally with low-order n-gram
  models.
\newblock In \emph{proceedings of the conference on Empirical Methods in
  Natural Language Processing}, pp.\  812--819. Association for Computational
  Linguistics, 2008.

\bibitem[Rumelhart et~al.(1986)Rumelhart, Hinton, and Williams]{backprop}
Rumelhart, David~E., Hinton, Geoffrey~E., and Williams, Ronald~J.
\newblock Learning representations by back-propagating errors.
\newblock \emph{Nature}, 323:\penalty0 533--536, 1986.

\bibitem[Snyder et~al.(2010)Snyder, Barzilay, and Knight]{snyder}
Snyder, Benjamin, Barzilay, Regina, and Knight, Kevin.
\newblock A statistical model for lost language decipherment.
\newblock In \emph{Proceedings of the 48th Annual Meeting of the Association
  for Computational Linguistics}, pp.\  1048--1057. Association for
  Computational Linguistics, 2010.

\bibitem[Tzeng et~al.(2014)Tzeng, Hoffman, Zhang, Saenko, and
  Darrell]{domain-confusion1}
Tzeng, Eric, Hoffman, Judy, Zhang, Ning, Saenko, Kate, and Darrell, Trevor.
\newblock Deep domain confusion: Maximizing for domain invariance.
\newblock \emph{arXiv preprint arXiv:1412.3474}, 2014.

\end{thebibliography}
\bibliographystyle{iclr2016_conference}

\end{document}